\documentclass[runningheads]{llncs}

\usepackage{eccv}

\usepackage{eccvabbrv}

\usepackage{graphicx}
\usepackage{booktabs}

\usepackage[accsupp]{axessibility}  

\usepackage[breaklinks,colorlinks,citecolor=eccvblue]{hyperref}

\usepackage{orcidlink}
\usepackage[T1]{fontenc}
\usepackage{amsmath}
\usepackage{amssymb}
\usepackage{pifont}
\usepackage{enumitem}
\usepackage{algorithm}
\usepackage{algpseudocode}
\usepackage{multirow}
\usepackage{makecell}
\usepackage{array}
\usepackage{hhline}
\usepackage{tabularx}
\usepackage{ragged2e}
\usepackage{nccmath}

\usepackage{subcaption}
\usepackage{caption}
\usepackage{tablefootnote}

\definecolor{lightcyan}{rgb}{0.88, 1.0, 1.0} 
\usepackage{graphicx}
\def\numImg{{K}}

\def\bboxSet{{\mathcal{B}}}
\def\bboxSetLidar{{\widehat{\bboxSet}}}
\def\imgSet{{\mathcal{I}}}
\def\pointCloud{{\mathcal{P}}}
\def\clsSet{{\Lambda}}
\def\clsSymbol{{\lambda}}
\newcommand{\img}[1]{I_{#1}}
\newcommand{\stereoPair}[1]{(\img{#1}^l, \img{#1}^r)}
\def\bboxSetImages{{\mathcal{R}}}
\newcommand{\bboxSetImg}[1]{\bboxSetImages_{#1}}

\def\TrMat{{T}}
\def\ProjMatrix{{P}}
\newcommand{\pMat}[1]{\ProjMatrix_{#1}}
\newcommand{\fMat}[2]{F_{#1#2}}

\newcommand{\cls}[1]{\clsSymbol_{#1}}
\def\yawAngle{{\theta}}
\newcommand{\yaw}[1]{\yawAngle_{#1}}
\newcommand{\bboxGeneric}[1]{b^{#1}}
\newcommand{\bboxLidar}[1]{b_{#1}^{3d}}
\newcommand{\bboxImg}[1]{b_{#1}^{2d}}

\newcommand{\real}[1]{\mathbb{R}^{#1}}

\def\Matching{{\mathcal{M}}}
\def\MatchingRecover{{\mathcal{A}}}
\def\MatchingFinal{{\mathcal{D}}}
\newcommand{\MatchingSingle}[2]{\Matching_{#1}^{#2}}
\newcommand{\MatchingRecoverSingle}[2]{\MatchingRecover_{#1}^{#2}}
\newcommand{\MatchingFinalSingle}[2]{\MatchingFinal_{#1}^{#2}}
\def\bboxSetImagesUnmatch{{\mathcal{U}}}
\newcommand{\bboxSetImgUnmatch}[2]{\bboxSetImagesUnmatch_{#1}^{#2}}

\def\varSymbol{{x}}

\def\prior{{p(y)}}
\newcommand{\posteriorMod}[1]{p(y|x_{#1})}
\newcommand{\posterior}[1]{p(y|\{x_i\}_{i=1}^{#1})}

\def\cornersSetLidar{{C}} 

\DeclareMathOperator*{\argmax}{argmax}

\def\iouThrBBoxMatching{\tau_b}
\def\iouThrDetRec{\tau_r}
\def\minPointFrustum{p_{\min}}
\def\enlargeFactor{e}

\begin{document}
\title{A Multimodal Hybrid Late-Cascade Fusion Network for Enhanced 3D Object Detection}
%
%
\author{Carlo Sgaravatti \and
Roberto Basla \and Riccardo Pieroni \and Matteo Corno
\and Sergio M. Savaresi \and Luca Magri \and Giacomo Boracchi}
\authorrunning{C. Sgaravatti et al.}
\titlerunning{A Multimodal Hybrid Late-Cascade Fusion Network}
%
\institute{Politecnico di Milano, Italy} 
%
\maketitle              
%

\begin{abstract}
We present a new way to detect 3D objects from multimodal inputs, leveraging both LiDAR and RGB cameras in a hybrid late-cascade scheme, that combines an RGB detection network and a 3D LiDAR detector. We exploit late fusion principles to reduce LiDAR False Positives, matching LiDAR detections with RGB ones by projecting the LiDAR bounding boxes on the image. We rely on cascade fusion principles to recover LiDAR False Negatives leveraging epipolar constraints and frustums generated by RGB detections of separate views. Our solution can be plugged on top of any underlying single-modal detectors, enabling a flexible training process that can take advantage of pre-trained LiDAR and RGB detectors, or train the two branches separately. We evaluate our results on the KITTI object detection benchmark, showing significant performance improvements, especially for the detection of Pedestrians and Cyclists. Code can be downloaded from:\\\href{https://github.com/CarloSgaravatti/HybridLateCascadeFusion}{https://github.com/CarloSgaravatti/HybridLateCascadeFusion}.

\end{abstract}

\begin{figure}[t]
\includegraphics[width=\textwidth]{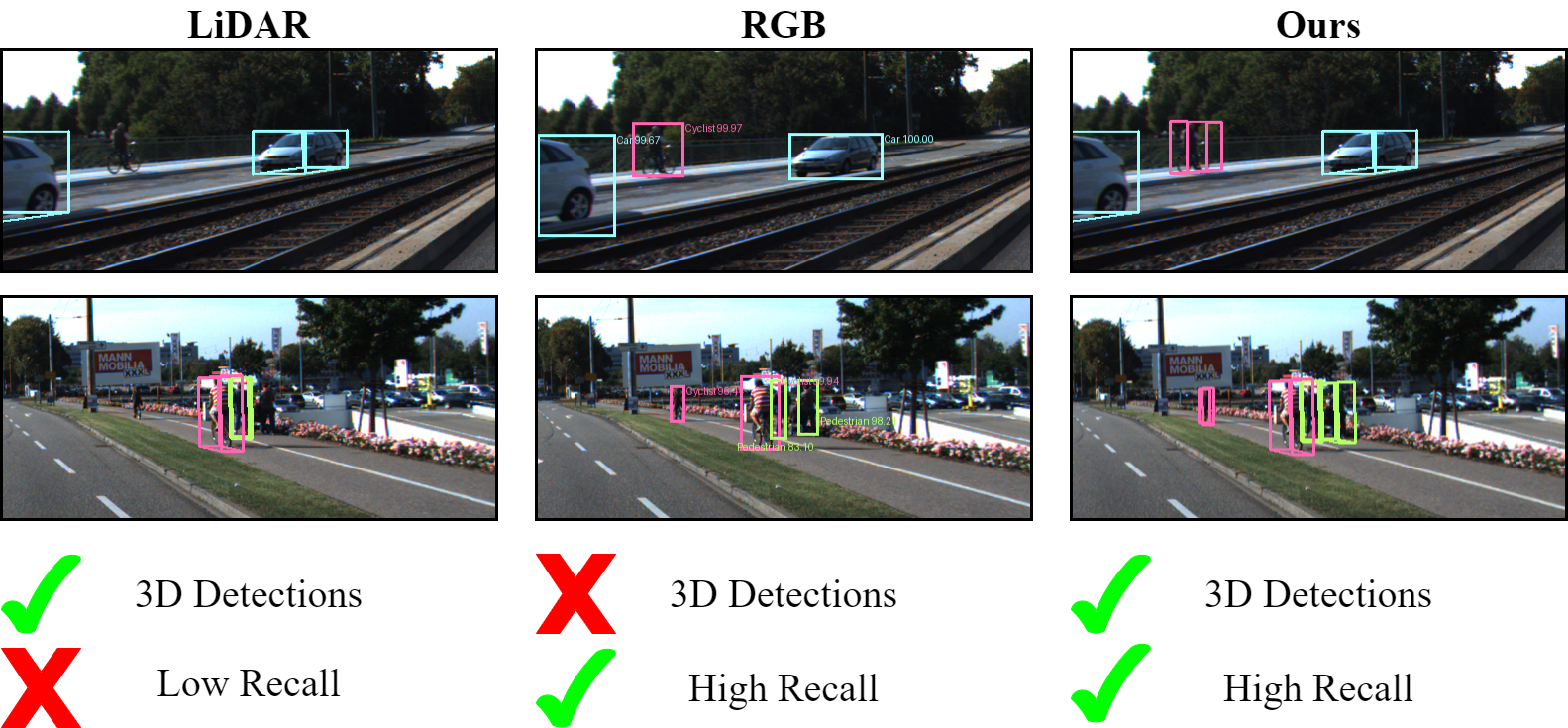}
\caption{\textbf{Left:} LiDAR branch struggles in detecting cyclists and pedestrians. \textbf{Center:} RGB branch correctly detects all the objects but lacks 3D information. \textbf{Right:} Our method recovers all the detections missed by the LiDAR and provides 3D information.}
\label{fig:teaser-image}
\end{figure}

\section{Introduction}

3D Object Detection is a fundamental task in Computer Vision. The goal is to locate objects in  3D starting from 3D measurements and/or RGB images. 3D Object Detection solutions are broadly applied in Autonomous Vehicles (AV) where finding the location, dimension and orientation of cars, pedestrians and cyclists is key for safe navigation and road safety \cite{kitti_dataset,cordts2016cityscapes,qian20223d}. Solutions based on Deep Neural Networks can be  \emph{single-modal}~\cite{second,pointpillars,std,pv-rcnn,se-ssd} or \emph{multi-modal}~\cite{mv3d,liu2023bevfusion,sparse-fuse-dense,frustum-pointnet,clocs_late_fusion}. Single-modal detectors usually process either RGB images or LiDAR Point Clouds, while multimodal ones improve the accuracy of 3D Object Detection thanks to complementary information sources \cite{survey_mm_autonomous_drive,mao20233d}. Point Clouds allow an accurate representation of the 3D scene's geometry, but their sparsity does not permit a full understanding of the semantics of the scene. Indeed, LiDAR-based detectors can accurately detect cars, but they struggle to detect occluded, small or distant objects \cite{qian20223d}. Unlike Point Clouds, RGB images do not provide depth and single-modal 3D detectors from RGB images struggle to accurately localize objects in 3D \cite{qian20223d}. In contrast, RGB images do provide rich semantic information that can be used to distinguish small objects such as cyclists and pedestrians, especially when they are far from the sensor. \cref{fig:teaser-image} highlights the main strengths of our method. The LiDAR branch struggles with challenging objects like cyclists and pedestrians that are not detected in both examples. Our method can provide a higher recall by recovering such missed 3D objects leveraging RGB 2D information.

The main difficulty in training a multimodal object detection network is that LiDAR and RGB images have completely different data representations, namely a scattered set of 3D points and a fixed-size tensor. Based on how these representations are fused, multimodal approaches can be classified into \emph{early} fusion, \emph{cascade} fusion and \emph{late} fusion \cite{survey_mm_autonomous_drive}. Early fusion  \cite{mv3d,avod,sparse-fuse-dense} combines the two information sources in the first stages of an end-to-end trainable network. The fusion of rich intermediate features comes at the cost of requiring paired data for training, and results in additional computational overhead, critical for real-time applications \cite{mao20233d}. Cascade fusion~\cite{frustum-pointnet,frustum-convnet} exploits a 2D RGB detector to find region proposals in the 3D space, thus they heavily suffer limitations of RGB detectors. Finally, late fusion methods exploit two parallel RGB and LiDAR branches, focusing on filtering out False Positive LiDAR detections \cite{towards_long_tailed,long_tailed_late_fusion}.  

To improve the detection accuracy for challenging classes like Cyclists and Pedestrians, we propose a hybrid late-cascade fusion approach. The proposed solution is illustrated in \cref{fig:architecture} and combines state-of-the-art methods for 3D and 2D detection in the context of a stereo camera system. Our key intuition is to recover missed detections of small or distant objects from the LiDAR branch by leveraging the geometric constraints between 3D and 2D predictions of LiDAR and RGB branches and between 2D predictions of different views in the RGB branch. We exploit late fusion principles to match the detections predicted by the two single-modal branches and filter our LiDAR False Positives. We take advantage of both the computational efficiency of single-modal detectors and the possibilities of training the two networks separately or using pre-trained models. Moreover, we rely on cascade fusion principles \cite{frustum-pointnet,frustum-convnet} to recover LiDAR False Negatives by extracting specific regions of the 3D space from RGB detections that are not associated with any LiDAR detection. Specifically, we first exploit geometric consistencies between 3D and 2D detections, to confirm or filter out LiDAR detections. In particular, we project 3D bounding boxes on the images and match these with 2D detections by solving an optimization problem based on the Intersection over Union (IoU). Then, to retrieve missed objects in 3D, we use epipolar constraints to pair 2D detections from the two images, and then we intersect their frustums. The point cloud at the frustums' intersection is fed to a specific 3D localization model to estimate the 3D bounding box.
Our solution is based on simple rules and lightweight detection modules, that add on top of existing --- possibly pre-trained --- image and LiDAR detection networks. Our approach does not incur in high computational overheads since we exploit the cascade principle only for regions where the LiDAR detector fails.

Our contribution is two-fold:\\
\emph{i}) We propose a novel hybrid fusion solution, combining late and cascade fusion approaches, to deal with Multimodal 3D Object Detection.\\
\emph{ii}) We design a Detection Recovery module, exploiting an epipolar-based assignment procedure to assign pairs of 2D detections from different views.\\
Our solution, evaluated on the KITTI benchmark \cite{kitti_dataset}, improves single-modal LiDAR detectors, especially for cyclists and pedestrians,  outperforming in some categories also multimodal detectors based on ad-hoc architectures. Extensive ablation studies demonstrate the effectiveness of our approach.

\section{Related Work}
3D Object Detection networks for AV can be divided into RGB-based, LiDAR-based and Multi-modal detectors. 

\textbf{RGB Detectors}. Despite the lack of depth information and the presence of occlusions that characterize RGB images, it is possible to train RGB-based object detection networks to extract 3D information from images \cite{pseudo-lidar-depth,wang2021fcos3d,monocular,3d-bbox-estimation-deep-learning-geometry}. These approaches have gained a lot of attention due to the low cost of the camera and the maturity of CNNs for extracting features from images. However, while 3D RGB detectors can successfully leverage the semantics of the image representation, they are usually characterized by poor 3D localization accuracy.

\textbf{LiDAR-based Detectors}.
Object detection is more challenging on 3D Point Clouds rather than in images, due to their sparse and scattered nature.
Several Deep Learning architectures have been proposed to address this challenge for LiDAR Point Clouds.  In particular, Point-based methods \cite{point-rcnn,std,3d-ssd} extract features by applying point operators to the raw point cloud, while Voxel-based approaches \cite{voxelnet,second,cia-ssd,se-ssd,partA2,transformation-equivariant} encode the point cloud into voxels and apply 3D CNNs to extract features. PointPillars \cite{pointpillars} extracts features on vertical columns and projects them into the Bird's Eye View (BEV) before applying 2D CNNs. PV-RCNN \cite{pv-rcnn} exploits the advantages of both voxel-based and point-based approaches defining a Voxel Set Abstraction module to integrate voxel features into key points sampled from the raw point cloud. While LiDAR-based detectors accurately detect objects like Cars, the sparsity of the point cloud does not allow for the same precision in detecting smaller objects like Pedestrians and Cyclists.

\textbf{Multimodal 3D Object Detection}. Multimodal approaches for 3D Object Detection have gained a lot of popularity over the last few years. These methods can be divided into three categories, depending on the processing stage in which the RGB and LiDAR data are fused \cite{survey_mm_autonomous_drive}.

Early fusion solutions usually combine the features from two modalities in the early stage of an end-to-end trainable network. Prominent examples are: MV3D \cite{mv3d}, which builds 3D proposals from the BEV and extracts RoI features from the images using the proposals projections, AVOD \cite{avod} and BEVFusion \cite{liu2023bevfusion}, that fuse the image features with the LiDAR ones on the BEV, PointFusion \cite{point-fusion}, that projects 3D points in the image and concatenates RGB features of the corresponding pixels to the features of the points, SFD \cite{sparse-fuse-dense} and VirConv \cite{virconv}, that use Depth Completion to build a dense pseudo point cloud from the image to be fused with the original point cloud. While early fusion networks have shown promising results, they are limited by the shortage of large-scale multimodal datasets \cite{survey_mm_autonomous_drive}. Indeed, the application of Data Augmentation, usually employed to solve data scarcity issues \cite{perez2017effectiveness,data-aug-3d,Zhou2024ASO}, to multimodal data is limited by the necessity of maintaining alignment between the two modalities \cite{survey_mm_autonomous_drive}.

Cascade fusion approaches first process the RGB data to produce either bounding boxes or segmentation masks and use these to enrich or crop the raw point cloud. PointPainting \cite{point-painting} makes use of a 2D semantic segmentation network to enrich the point cloud with segmentation masks. Faraway-Frustum \cite{faraway_frustum} combines a MaskRCNN and Frustum Network to detect objects that are far from the sensor. FrustumPointNet \cite{frustum-pointnet}, FrustumConvNet \cite{frustum-convnet} and FrustumPointPillars \cite{frustum_pointpillars} lift 2D RGB detections to frustums to reduce the 3D search space for the LiDAR detector. However, cascade fusion solutions are limited by the performance of 2D detectors. Our approach exploits cascade fusion principles, but we first rely on the LiDAR detector to find a set of objects from the scene geometry. Then, we pair RGB detections to find missed 3D detections.

Late fusion approaches leverage two parallel object detection networks from each modality and combine their outputs in a final module. CLOCs \cite{clocs_late_fusion} exploits Geometric and Semantic consistency between LiDAR 3D detections and RGB 2D detections and builds a fusion network to adjust the confidence scores of the 3D detections. Çaldıran \etal \cite{late_fusion_with_segmentation} filter out LiDAR False Positive detections with an asymmetric late fusion approach. Recently, Peri \etal \cite{towards_long_tailed} use a 3D RGB detector to filter the LiDAR detections that are not near any RGB detection according to the distance between the bounding box centers. Differently, Ma \etal \cite{long_tailed_late_fusion} use a 2D RGB detector and match LiDAR detections with RGB ones on the image plane. These approaches allow removing LiDAR False Positive detections but assume a high recall for the LiDAR detector. Our work also recovers LiDAR False Negatives by exploiting unmatched RGB detections to find new objects.

\section{Problem Formulation}

The input of our multi-modal 3D Object Detector is a set of $\numImg$ pairs of stereo images $\imgSet = \{\stereoPair{1}, ..., \stereoPair{\numImg}\}$, where $\img{i}^l \in \real{W_i^l \times H_i^l \times 3}$ and $\img{i}^r \in \real{W_i^r \times H_i^r \times 3}$ correspond to views of the same scene from the left ($l$) and right ($r$) cameras, and a Point Cloud containing $M$ points $\pointCloud = \{p_1, p_2, ...,  p_M\}$, $p_j = (x_j, y_j, z_j, r_j)^T \in \real{4}$, where $(x_j, y_j, z_j)$ is the position of the point $p_j$ and $r_j$ is the corresponding reflectance. The point cloud is expressed in LiDAR coordinates, with $\TrMat$ being the known transformation matrix from LiDAR to camera coordinates, which are in the coordinate system of a reference camera, \eg $\img{1}^l$. We assume to know for each image $\img{i}^q$, with $q \in \{l,r\}$, the camera matrix $\pMat{i}^q \in \real{3 \times 4}$, that projects 3D points in the coordinates of the reference camera to the image plane.

The 3D Object Detection algorithm computes a set of 3D bounding boxes $\bboxSet$ surrounding the objects in the 3D space:
\begin{equation} \label{eq:problem_form}
(\imgSet, \pointCloud) \longmapsto \bboxSet = \{(\bboxLidar{p}, s_p, \cls{p})  | \bboxLidar{p} \in \real{7}, s_p \in [0, 1], \cls{p} \in \clsSet, p = 1, \dots, P \}
\end{equation}
where $\bboxLidar{p} = (x_p, y_p, z_p, l_p, h_p, w_p, \yaw{p})^T$ concatenates the 3D coordinates of the center $(x_p, y_p, z_p)$, the dimensions $(l_p, h_p, w_p)$ of the bounding box, and $\yaw{p}$ is the yaw angle; $s_p$ is the confidence score, $\cls{p}$ is the associated label from the set of classes $\clsSet$ and $P$ is the number of detections.

\section{Proposed Solution}

\begin{figure}[t]
\centering
\includegraphics[width=0.9\textwidth]{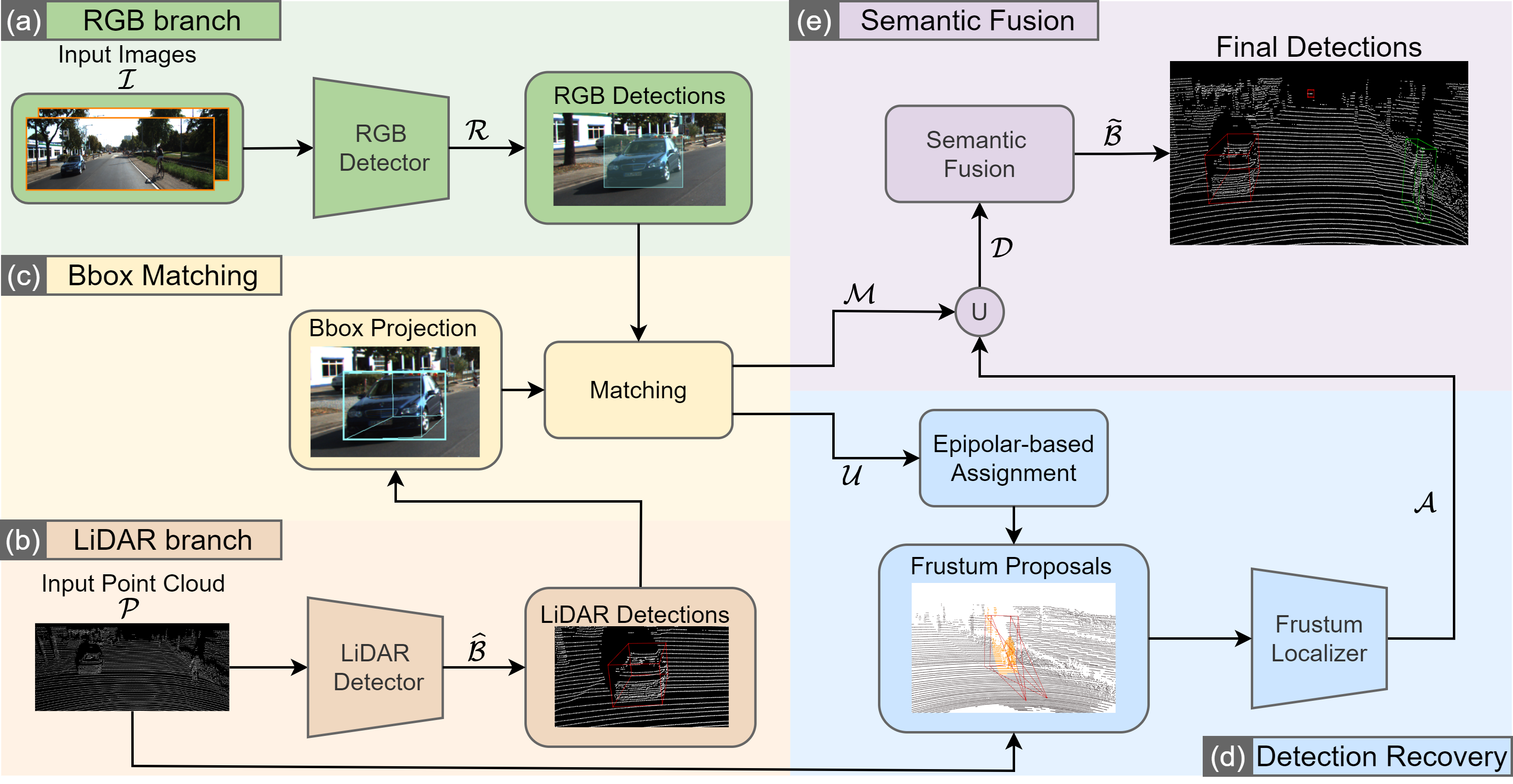}
\caption{Architecture. (a) the RGB branch outputs 2D detections $\bboxSetImages$ from each image. (b) the LiDAR branch  computes 3D detections $\bboxSetLidar$  from the input Point Cloud $\pointCloud$. (c) Bbox Matching projects the 3D detections and matches them with the 2D ones in each image ($\Matching$). (d) the unmatched RGB detections $\bboxSetImagesUnmatch$ are fed to the Detection Recovery module that matches 2D detections across stereo views, then extracts frustum proposals and uses the matched pairs to recover missed LiDAR detections ($\MatchingRecover$). (f) the Semantic Fusion module enforces semantic consistency between the LiDAR and the RGB branches.}
\label{fig:architecture}
\end{figure}

At a high level, our method comprises 5 modules, as depicted in \cref{fig:architecture}: RGB branch (a), LiDAR branch (b), Bbox Matching (c), Detection Recovery (d) and Semantic Fusion (e). The RGB and LiDAR branches leverage pre-trained models to predict a set of 2D and 3D detections, respectively. In the Bbox Matching module, 3D detections from the LiDAR branch are projected in every image and compared with 2D detections from the RGB branch. We compute the IoU between them to establish matches by solving an optimization problem. Unmatched RGB detections are fed to the Detection Recovery module where we compute Frustum Proposals to crop portions of Point Clouds that we further process to detect missed 3D objects. Finally, the Semantic Fusion module combines the labels in case the 2D and 3D predictions are discordant.

\subsection{Bounding Box Matching} \label{sec:bbox_matching}

\begin{algorithm}[t]
\scriptsize
\caption{Bbox Matching}\label{alg:matching}
\hspace*{\algorithmicindent} \textbf{Input}:The set of RGB detections $\bboxSetImages$, the LiDAR detections $\bboxSetLidar$, the calibration matrices $(\TrMat, \{(\pMat{i}^l, \pMat{i}^r)\}_{i=1}^\numImg)$ and the IoU threshold $\iouThrBBoxMatching$ \\
\hspace*{\algorithmicindent} \textbf{Output}: Matched pairs and unmatched RGB detections $(\Matching, \bboxSetImagesUnmatch)$
\begin{algorithmic}[1]
\Function{BboxMatching}{$\bboxSetImages$, $\bboxSetLidar$, $\TrMat$, $\{(\pMat{i}^l, \pMat{i}^r)\}_{i=1}^\numImg$, $\iouThrBBoxMatching$}
\State $\Matching, \bboxSetImagesUnmatch \gets \emptyset, \emptyset$
\State $\cornersSetLidar \gets \Call{ExtractCorners}{\bboxSetLidar}$
\State $\cornersSetLidar \gets \Call{TransformCoordinates}{\cornersSetLidar, \TrMat}$
\For{\texttt{$(i, q) \in \{1, \dots, \numImg\} \times \{l,r\}$}}
\State $\cornersSetLidar_i^q \gets \Call{ProjectCorners}{\cornersSetLidar, P_i^q}$
\State $\bboxSet_i^q \gets \Call{AxisAlignedBboxes}{\cornersSetLidar_i^q}$
\State $(\MatchingSingle{i}{q}, \bboxSetImgUnmatch{i}{q}) \gets \Call{IouAssignment}{\bboxSet_i^q, \bboxSetImg{i}^q, \iouThrBBoxMatching}$
\State $\Matching \gets \Matching \cup \{\MatchingSingle{i}{q}\}$
\State $\bboxSetImagesUnmatch \gets \bboxSetImagesUnmatch \cup \{\bboxSetImgUnmatch{i}{q}\}$
\EndFor
\State \Return $(\Matching, \bboxSetImagesUnmatch)$
\EndFunction
\end{algorithmic}
\end{algorithm}

Let us assume the LiDAR branch returns a collection of 3D bounding boxes $\bboxSetLidar$, as in \cref{eq:problem_form}. 
Similarly, the RGB branch predicts a set of 2D bounding boxes $\bboxSetImg{i}^q = \{(\bboxImg{r}, s_r, \cls{r})\}_{r=1}^R$, for each single image $\img{i}^q$. 
Our Bbox Matching module aims at matching every $\bboxLidar{p}$ in $\bboxSetLidar$ with possibly a single $\bboxImg{r}$.  Once the 3D bounding boxes are projected in the image planes, this boils down to solving an assignment problem that maximizes their IoU with  $\bboxImg{r}$. Specifically, as detailed in \cref{alg:matching}, we extract corners $\{p_1, \dots,  p_8\}$ of each $\bboxLidar{p}$, expressed in homogeneous coordinates in the LiDAR reference system (Line 4). Then, we move the corners in the world reference system $\TrMat p_j$ and project them to $\widetilde{p}_j = \pMat{i}^q\TrMat p_j$ using the camera matrix $\pMat{i}^q$ for each image $\img{i}^q$. We then extract an axis-aligned 2d bounding box $\bboxGeneric{proj}_p$ enclosing the projected corners (Lines 6-7). The assignment problem then becomes:

\begin{subequations}\label{eq:linear_sum_assign}
\noindent
\begin{minipage}[b]{0.07\textwidth}
    $\max_{\varSymbol}$
\end{minipage}
\begin{minipage}[b]{0.425\textwidth}
    \begin{fleqn}
    \begin{equation}
    \sum\limits_{p,r} IoU(\bboxGeneric{proj}_p, \bboxImg{r}) \varSymbol_{pr} \tag{\ref{eq:linear_sum_assign}}
    \end{equation}
    \end{fleqn}
\end{minipage}
\begin{minipage}[b]{0.06\textwidth}
    \hfill \text{s.t.}
\end{minipage}
\begin{minipage}[b]{0.425\textwidth}
    \begin{fleqn}
    \begin{equation}
    \sum\limits_{p,r} \varSymbol_{pr} \ge \min\{P, R\}\label{eq:linear_sum_assign:constr3}
    \end{equation}
    \end{fleqn}
\end{minipage}\\
\noindent
\begin{minipage}[b]{0.07\textwidth}
\hfill
\end{minipage}
\begin{minipage}[b]{0.425\textwidth}
    \begin{fleqn}
    \begin{equation}
    \sum\limits_p \varSymbol_{pr} \leq 1, \hspace{0.25em} \forall r \in \{1, \dots, R\} \label{eq:linear_sum_assign:constr2}
    \end{equation}
    \end{fleqn}
\end{minipage}
\begin{minipage}[b]{0.05\textwidth}
\hfill
\end{minipage}
\begin{minipage}[b]{0.425\textwidth}
    \begin{fleqn}
    \begin{equation}
    \sum\limits_r \varSymbol_{pr} \leq 1, \hspace{0.25em} \forall p \in \{1, \dots, P\} \label{eq:linear_sum_assign:constr1}
    \end{equation}
    \end{fleqn}
\end{minipage}
\end{subequations}
where $\varSymbol_{pr} \in \{0,1\}$ denotes if $\bboxLidar{p}$ and $\bboxImg{r}$ are matched or not. The constraints \eqref{eq:linear_sum_assign:constr1} and \eqref{eq:linear_sum_assign:constr2} specify that each detection in one image should be assigned with at most one detection in the other image, while \eqref{eq:linear_sum_assign:constr3} enforces the assignment of all instances in a set. We solve the optimization problem with the Jonker-Volgenant algorithm \cite{lsap-alg}. We denote with $\Matching$ the set of matched bounding boxes. We consider 3D detections that have no matches in any image as False Positives (FP) of the LiDAR branch and we remove them.

Since our Bbox Matching procedure prunes out irrelevant 3D detections, we can adjust thresholds of the LiDAR branch and in particular we \emph{i}) lower the threshold on the confidence score to include more detections in $\bboxSetLidar$, and \emph{ii}) relax the IoU threshold in Non Maxima Suppression (NMS) to increase the number of 3D bounding boxes considered. We set both these thresholds to 0.3. As regards the threshold on the confidence score for the RGB branch, we fix it to 0.5 to guarantee the overall precision on 2D detections, so irrelevant LiDAR detections will not be matched. Finally, once the matching is performed, we remove from $\Matching$ all the matches with an IoU below $\iouThrBBoxMatching$ and add the corresponding RGB detections to the set of unmatched RGB detections $\bboxSetImagesUnmatch$, which will be processed in the Detection Recovery module (\cref{sec:det_recover}). Please, note that the matching procedure is performed using only the geometry of the output detections and the 3D stereo vision constraints; we do not enforce semantic consistency at this level since LiDAR and RGB detections may not predict the same semantic class.


\subsection{Detection Recovery} \label{sec:det_recover}

The set of unmatched RGB detections $\bboxSetImagesUnmatch$ typically corresponds to small and/or distant objects that the LiDAR branch has missed. Starting from $\bboxSetImagesUnmatch$, our Detection Recovery module aims to recover the corresponding missed 3D detections by leveraging two-view geometry of a stereo pair $(\img{i}^l, \img{i}^r)$. The 
output is a set $\MatchingRecover = \{\MatchingRecoverSingle{i}{q} | i \in \{1, \dots, \numImg\}, q \in \{l, r\}\}$ of pairs of RGB-LiDAR detections:
\begin{equation}
    \MatchingRecoverSingle{i}{q} := \{(\bboxLidar{j}, \bboxImg{j}, s_j^{3d}, s_j^{2d}, \cls{j}^{3d}, \cls{j}^{2d}) | (\bboxImg{j}, s_j^{2d}, \cls{j}^{2d}) \in \bboxSetImg{i}^q\},
\end{equation}
where $(\bboxLidar{j}, s_j^{3d}, \cls{j}^{3d})$ are the new 3D detections recovered.

At a high level, as detailed in \cref{alg:recovery}, the recovery of 3D detections is performed in 3 steps. First, each bounding box $\bboxImg{l} \in \bboxSetImgUnmatch{j}{l}$ in the left image is possibly matched to a corresponding bounding box $\bboxImg{r} \in \bboxSetImgUnmatch{j}{r}$ in the right view (Lines 3-5). Second, each pair of matched bounding boxes is backprojected to crop a 3D region (Line 7). Third, an ad-hoc Frustrum Localizer is used to detect objects in the cropped 3D region. 3D predictions are then validated by checking the geometric consistency with the input images (Lines 9-15).

\begin{algorithm}[t]
\scriptsize
\caption{Detection Recovery}\label{alg:recovery}
\hspace*{\algorithmicindent} \textbf{Input}: Unmatched detections $(\bboxSetImgUnmatch{i}{l}, \bboxSetImgUnmatch{i}{r})$, the Point Cloud $\pointCloud$, the calibration matrices $(\TrMat, \pMat{i}^l, \pMat{i}^r)$, the minimum number of points $\minPointFrustum$, the IoU threshold $\iouThrDetRec$ and the enlargement factor $\enlargeFactor$ \\
\hspace*{\algorithmicindent} \textbf{Output}: Recovered pairs of detections $(\MatchingRecoverSingle{i}{l}, \MatchingRecoverSingle{i}{r})$
\begin{algorithmic}[1]
\Function{DetectionRecovery}{$\bboxSetImgUnmatch{i}{l}$, $\bboxSetImgUnmatch{i}{r}$, $\pointCloud$, $\TrMat$, $\pMat{i}^l$, $\pMat{i}^r$, $\minPointFrustum$, $\iouThrDetRec$, $\enlargeFactor$}
    \State $\MatchingRecoverSingle{i}{l}, \MatchingRecoverSingle{i}{r} \gets \emptyset, \emptyset$
    \State $\fMat{l}{r}^i \gets \Call{FundamentalMatrix}{\pMat{i}^l, \pMat{i}^r}$ \hfill \Comment{Epipolar-based assignment}
    \State $\mathcal{D} \gets \Call{ComputeDistanceMatrix}{\bboxSetImgUnmatch{i}{l}, \bboxSetImgUnmatch{i}{r}, \fMat{l}{r}^i}$
    \State $\mathcal{M}_{2d} \gets \Call{Assign}{D, \bboxSetImgUnmatch{i}{l}, \bboxSetImgUnmatch{i}{r}}$
    \For{\texttt{$(\bboxImg{l}, s_{l}^{2d}, \cls{l}^{2d}, \bboxImg{r}, s_{r}^{2d}, \cls{r}^{2d}) \in \mathcal{M}_{2d}$}}
        \State $\pointCloud_{lr} \gets \Call{CropPointCloud}{\pointCloud, \bboxImg{l}, \bboxImg{r}, \pMat{i}^l, \pMat{i}^r, \TrMat, \enlargeFactor}$ \hfill \Comment{Frustum Proposal}
        \If{$|\pointCloud_{lr}| > \minPointFrustum$}
            \State $\bboxGeneric{3d} \gets \Call{FrustumLocalizer}{\pointCloud_{lr}}$ \hfill \Comment{Localization}
            \State $\bboxGeneric{proj}_l, \bboxGeneric{proj}_r \gets \Call{BboxProjections}{\bboxGeneric{3d}, \TrMat, \pMat{i}^l, \pMat{i}^r}$
            \If{$\max(IoU(\bboxGeneric{proj}_l, \bboxImg{l}), IoU(\bboxGeneric{proj}_r, \bboxImg{r})) > \iouThrDetRec$} \hfill \Comment{Geometric Consistency}
                \State $s_{RGB}, \clsSymbol^{\prime} \gets \Call{MostConfidentRGB}{s_{l}^{2d}, \cls{l}^{2d}, s_{r}^{2d}, \cls{r}^{2d}}$
                \State $s^{\prime} \gets s_{RGB} \cdot IoU(\bboxGeneric{proj}_l, \bboxImg{l}) \cdot IoU(\bboxGeneric{proj}_r, \bboxImg{r})$
                \State $\MatchingRecoverSingle{i}{l} \gets \MatchingRecoverSingle{i}{l} \cup \{(\bboxGeneric{3d}, s^{\prime}, \clsSymbol^{\prime}, \bboxImg{l}, s_{l}^{2d}, \cls{l}^{2d})\}$
                \State $\MatchingRecoverSingle{i}{r} \gets \MatchingRecoverSingle{i}{r} \cup \{(\bboxGeneric{3d}, s^{\prime}, \clsSymbol^{\prime}, \bboxImg{r}, s_{r}^{2d}, \cls{r}^{2d})\}$
            \EndIf
        \EndIf
    \EndFor
    \State \Return $(\MatchingRecoverSingle{i}{l}, \MatchingRecoverSingle{i}{r})$
\EndFunction
\end{algorithmic}
\end{algorithm}

\begin{figure}[t]
    \centering
    \begin{minipage}{0.45\textwidth}
        \centering
        \begin{subfigure}[b]{\linewidth}
            \centering
            \includegraphics[width=\linewidth]{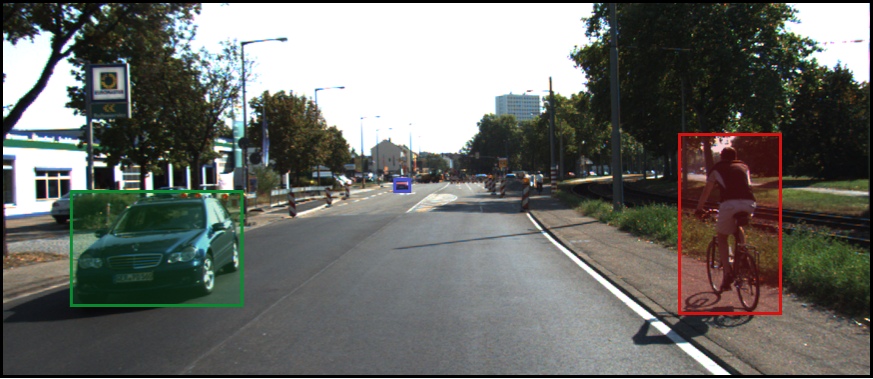}
            \caption{2D detections on the left image}
            \label{fig:det_recovery:left_bboxes}
        \end{subfigure}
        \vfill
        \begin{subfigure}[b]{\linewidth}
            \centering
            \includegraphics[width=\linewidth]{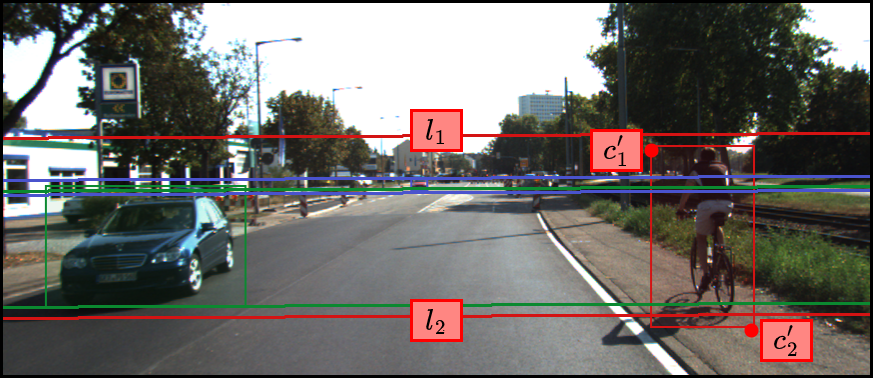}
            \caption{2D detections and epipolar lines on the right image}
            \label{fig:det_recovery:right_epipolar}
        \end{subfigure}
    \end{minipage}
    \hfill
    \begin{minipage}{0.45\textwidth}
        \begin{subfigure}[c]{\linewidth}
            \centering
            \includegraphics[width=\linewidth]{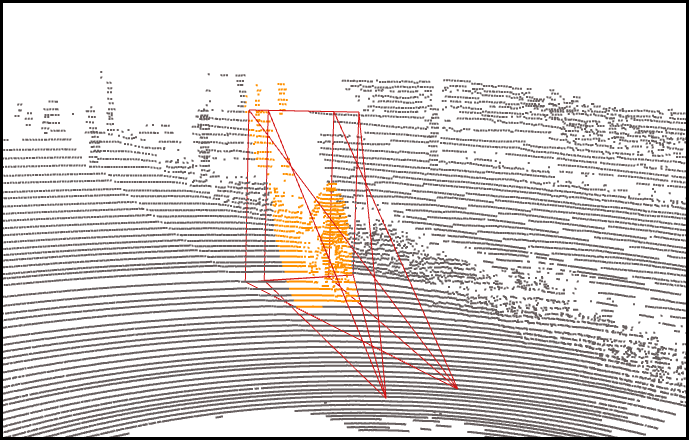}
            \caption{Frustum point cloud obtained by selecting the points inside the pair of frustums given by the two assigned detections.}
            \label{fig:det_recovery:frustum}
        \end{subfigure}
    \end{minipage}
    \caption{Illustration of the frustum proposals, obtained from the Detection Recovery module assignment procedure between two stereo images $\img{i}^l$ and $\img{i}^r$.}
    \label{fig:det_recovery}
\end{figure}

The left-right Bounding Box matching is cast as an assignment problem similar to \cref{eq:linear_sum_assign}, but instead of maximizing the IoU, here we minimize a distance between bounding boxes based on epipolar geometry.
Ideally, the corners of a bounding box in the right image should belong to the epipolar lines defined by the corners of the corresponding bounding box in the left image. When the stereo pair is rectified as in \cref{fig:det_recovery:right_epipolar}, the epipolar lines are horizontal.  However, the predictions of the RGB branch may have small inconsistencies, as shown in \cref{fig:det_recovery:left_bboxes,fig:det_recovery:right_epipolar},
but still the corners $(c_1^\prime, c_2^\prime)$ are expected to be close to the epipolar lines $(l_1, l_2)$ defined by the bounding box in the other image. This is illustrated for the bounding box of the cyclist in \cref{fig:det_recovery:right_epipolar}.
We thus define the cost  $d(\cdot,\cdot)$ for matching  $\bboxImg{l}$ and $\bboxImg{r}$ as the sum of the Euclidean distances $\tilde{d}(\cdot,\cdot)$ between each corner of $\bboxImg{r}$ and the epipolar lines of the corresponding corner of $\bboxImg{l}$ (Line~5):
\begin{equation} \label{eq:distance_epipolar}
    d(\bboxImg{l}, \bboxImg{r}) = \tilde{d}(l_1, c_1^\prime) + \tilde{d}(l_2, c_2^\prime).
\end{equation}
We solve the corresponding assignment problem to get matches $\mathcal{M}_{2d}$ using the Jonker-Volgenant algorithm. We then back-project and intersect in the 3D space each pair of bounding boxes in $\mathcal{M}_{2d}$, and we obtain Frustum Proposals (\cref{fig:det_recovery:frustum}) containing portions of LiDAR point cloud (Line 7). In practice, 2D detections may be slightly geometrically inaccurate, thus their Frustum Proposal might cut away useful points. Therefore, we enlarge the 2D bounding boxes by an enlargement factor $\enlargeFactor$ for width and height, keeping the centers of the bounding boxes fixed. Frustum Proposals containing fewer points than a certain threshold $\minPointFrustum$ are ignored; otherwise, each proposal is fed to the Frustum Localizer, which localizes the object in the 3D space. We adopt as Frustum Localizer FrustumPointNet \cite{frustum-pointnet}, and we enrich the Frustum Proposals input by a Gaussian mask proposed in FrustumPointPillars \cite{frustum_pointpillars}, added as an additional channel to the Point Cloud. We assign the estimated label and the score of the most confident RGB detection to the localized 3D object returned by the FrustumPointNet. However, since RGB detections are typically more confident than those in 3D, we down-weight the confidence score by the IoU with the 2D detections as (Line~13):
\begin{equation} \label{eq:iou-downweight}
    s^{3d} = s_{RGB} \cdot IoU(\bboxGeneric{proj}_l, \bboxImg{l}) \cdot IoU(\bboxGeneric{proj}_r, \bboxImg{r})
\end{equation}
where $s_{RGB}$ is the score extracted by the two stereo RGB detections, and $(\bboxGeneric{proj}_l, \bboxGeneric{proj}_r)$ are the projections in the two image planes of the bounding box predicted by the Frustum Localizer.
We also discard 3D objects having a projection on 2D bounding boxes with an IoU lower than a threshold $\iouThrDetRec$ (Lines~9-10).

We remark that all cascade fusion approaches based on frustums \cite{frustum-convnet,frustum-pointnet} have been designed for single-view settings. Our solution, leveraging stereo pairs, analyzes intersections of frustums from multiple views thus feeds the 3D localization network with selected points that most likely refer to the target object. Therefore, we expect the Detection Recovery module to better find challenging objects, \ie smaller or sparse objects.

\subsection{Semantic Fusion} \label{sec:semantic_fusion}

The Semantic Fusion module, detailed in \cref{alg:semantic_fusion}, enforces semantic consistency on all the 3D detection since matching RGB and LiDAR detections can refer to different predicted classes. In particular, the Semantic Fusion module replaces the LiDAR label and confidence score with the RGB ones \cite{long_tailed_late_fusion}, as we assume that RGB images contain better semantic information. The input of the semantic module contains the set of matched detections $\Matching$ from the Bbox Matching module and the set of recovered detections $\MatchingRecover$ from the Detection Recovery module, which we define as $\MatchingFinal = \{\MatchingFinalSingle{i}{q} | i \in \{1, \dots, \numImg\}, q \in \{l, r\}\}$, where $\MatchingFinalSingle{i}{q} = \MatchingSingle{i}{q} \cup \MatchingRecoverSingle{i}{q}$.   When multiple RGB views having different predicted classes are matched to the same 3D detection, we propagate the label from the most confident RGB detection (Lines 8-9). When all matching detections have the same predicted class, we adjust the confidence score of the LiDAR detections through the RGB confidence (Lines 10 and 13). We follow the probabilistic ensemble framework in \cite{chen2022multimodal}, which assumes conditional independence between different modalities, obtaining the following formulation of the final detection confidence score for class $y \in \clsSet$ with $L$ matching modalities\footnote{In our case, between LiDAR and one or two images depending on whether there is a match on both the stereo images or on only one of the two.}:
\begin{equation}
    \posterior{L} \propto \frac{\prod_{i=1}^L \posteriorMod{i}}{\prior^{L-1}}
\end{equation}
where $\posteriorMod{i}$ is the confidence score for the $i$-th matching modality, and $\prior$ is the class prior, which can be obtained by computing the per-class frequencies or treated as a uniform prior. In this work, we follow this second approach.

\begin{algorithm}[t]
\scriptsize
\caption{Semantic Fusion} \label{alg:semantic_fusion}
\hspace*{\algorithmicindent} \textbf{Input}: Matching detections in each view $\MatchingFinal$\\
\hspace*{\algorithmicindent} \textbf{Output}: Final detection output $\tilde{\bboxSet}$
\begin{algorithmic}[1]
\Function{SemanticFusion}{$\MatchingFinal$}
\State $\tilde{\bboxSet} \gets \emptyset$
\For{\texttt{$i = 1, \dots, \numImg$}}
    \State $\bboxSetLidar^i \gets \Call{GetUniqueLidarDetections}{\MatchingFinalSingle{i}{l}, \MatchingFinalSingle{i}{r}}$
    \For{\texttt{$(\bboxLidar{j}, s_j^{3d}, \cls{j}^{3d}) \in \bboxSetLidar^i$}}
        \If{$\Call{BothMatched}{\MatchingFinalSingle{i}{l}, \MatchingFinalSingle{i}{r}, \bboxLidar{j}}$}
            \State $(s_l^{2d}, \cls{l}^{2d}, s_r^{2d}, \cls{r}^{2d}) \gets \Call{GetMatchedSemantics}{\MatchingFinalSingle{i}{l}, \MatchingFinalSingle{i}{r}, \bboxLidar{j}}$
            \State $q_{max} \gets \argmax \{s_l^{2d}, s_r^{2d}\} $
            \State $\cls{j}^{\prime} \gets \clsSymbol_{q_{max}}^{2d}$
            \State $s_j^{\prime} \gets \Call{ProbabilisticEnsemble}{s_j^{3d}, \cls{j}^{3d}, s_l^{2d}, \cls{l}^{2d}, s_r^{2d}, \cls{r}^{2d}}$
        \Else
            \State $(s^{2d}, \cls{j}^{\prime}) \gets \Call{GetSingleMatchedSemantic}{\MatchingFinalSingle{i}{l}, \MatchingFinalSingle{i}{r}, \bboxLidar{j}}$
            \State $s_j^{\prime} \gets \Call{ProbabilisticEnsemble}{s_j^{3d}, \cls{j}^{3d}, s^{2d}, \cls{j}^{\prime}}$
        \EndIf
        \State $\tilde{\bboxSet} \gets \tilde{\bboxSet} \cup (\bboxLidar{j}, s_j^{\prime}, \cls{j}^{\prime})$
    \EndFor
\EndFor
\State \Return $\tilde{\bboxSet}$
\EndFunction
\end{algorithmic}
\end{algorithm}

\section{Experiments}

We evaluate our proposed solution on the KITTI object detection dataset \cite{kitti_dataset} and compare it against state-of-the-art single-modal (LiDAR only) and multi-modal detectors. Finally, we extensively ablate the components of our approach to demonstrate their effectiveness.

\subsection{Experimental Setup}

\textbf{Dataset}. The KITTI object detection \cite{kitti_dataset} dataset provides 7481 training samples and 7518 testing samples, with both LiDAR Point Clouds and RGB camera images. We follow the evaluation protocol defined in \cite{mv3d} to split the training dataset into 3712 training samples and 3769 validation samples and the KITTI evaluation protocol, which defines three classes of difficulties: easy, moderate and hard. Further details are in \cite{kitti_dataset}. We evaluate our approach using the 3D Average Precision (AP) and the BEV AP.

\medskip\noindent
\textbf{LiDAR/RGB Detectors}. We test our method using different LiDAR detectors: SECOND \cite{second}, PointPillars \cite{pointpillars}, PV-RCNN \cite{pv-rcnn} and PartA2 \cite{partA2}, from the MMDetection3D \cite{mmdet3d2020} framework. We use the pre-trained PointPillars, PV-RCNN and PartA2 models freely available from MMDetection3D. Differently, we train SECOND on the Point Clouds of the KITTI training set, using the parameters suggested by \cite{mmdet3d2020}, applying object noise, random flip on the BEV and ground-truth sampling as data augmentation procedures, and selecting the model associated with the highest 3D AP on the validation set at the 80th epoch, with 10 epochs as patience. As a 2D detector, we use a FasterRCNN \cite{faster_rcnn} using MMDetection's \cite{mmdetection} implementation, using ResNet101 \cite{resnet} as the backbone and a Feature Pyramid Network (FPN) \cite{fpn} as the neck to detect objects at different scales. We train the Faster RCNN model on the left images of the KITTI training set, applying data augmentation techniques from \cite{albu} to add Gaussian noise, motion blur and several transformations to simulate different climate conditions such as rain or sun flares. We use the 2D AP on the validation set to select the best model at the 200th epoch. To increase the performance on hard cases in all 2D detections and to prevent filtering out overlapped bounding boxes (due to occluded objects), we exploit Soft-NMS \cite{soft-nms}. For the Frustum Localizer, we re-implement Frustum PointNet \cite{frustum-pointnet} and train it to localize the objects on the cropped Point Clouds extracted by the KITTI training dataset RGB ground truths. As suggested in \cite{frustum-pointnet}, we add noise to the ground truth 2D bounding boxes to simulate inconsistencies. All the experiments were conducted on a cluster with multi-GPU nodes equipped with 8 A100.

\subsection{Performance Comparison with Existing Solutions}

We evaluate the performance of our late-cascade fusion module on the KITTI validation set, comparing it against single-modal LiDAR detectors and multi-modal frameworks. \cref{table:single_mod_comp_3d,table:single_mod_comp_bev} show, respectively, the 3D AP and BEV AP for Pedestrians, Cyclists and Cars. There are only marginal improvements for cars, for which the performance of LiDAR detectors is known to be good. In contrast, our method significantly increases the performance of single-modal detectors for pedestrians and cyclists. Specifically, since LiDAR-based detectors struggle to detect cyclists in the moderate and hard cases, in these two cases our method significantly improves the Cyclists' performance. As regards pedestrians, our method provides big improvements in all scenarios. \cref{table:multi_modal_comparison} compares the results of our method with multi-modal solutions on the KITTI validation set, showing how plugging PV-RCNN and Faster RCNN in our hybrid late-cascade framework permits reaching state-of-the-art results on pedestrian and cyclists. Moreover, by using PointPillars, we can provide competitive results with a lower computational time compared to current multi-modal solutions. Note that Frames Per Second (FPS) are taken from the corresponding original publications, thus the comparison of our solution is not carried out on identical computing architectures. However, our experiments indicate that the results are in line with our implementations. 

\begin{table}[t]
\centering
\caption{Comparison with single modal detectors (3D AP) on the KITTI val set.}
\resizebox{.92\textwidth}{!}{%
\begin{tabular}{|c|>{\centering\arraybackslash}p{1.2cm}|>{\centering\arraybackslash}p{1.2cm}|>{\centering\arraybackslash}p{1.2cm}|>{\centering\arraybackslash}p{1.2cm}|>{\centering\arraybackslash}p{1.2cm}|>{\centering\arraybackslash}p{1.2cm}|>{\centering\arraybackslash}p{1.2cm}|>{\centering\arraybackslash}p{1.2cm}|>{\centering\arraybackslash}p{1.2cm}|}
\hline
\multirow{2}{*}{Detector} & \multicolumn{3}{c|}{Car $AP_{3d}$} & \multicolumn{3}{c|}{Pedestrian $AP_{3d}$} & \multicolumn{3}{c|}{Cyclist $AP_{3d}$} \\
\cline{2-10}
& Easy & Mod. & Hard & Easy & Mod. & Hard & Easy & Mod. & Hard \\
\hhline{|=|=|=|=|=|=|=|=|=|=|}
SECOND & 87.83 & 78.46 & 73.75 & 59.12 & 52.78 & 47.41 & 75.58 & 61.73 & 58.18 \\
\hline
SECOND+FasterRCNN & 87.98 & 79.27 & 74.37 & 65.98 & 59.73 & 53.47 & 85.24 & 72.77 & 68.27 \\
\hline
\rowcolor{lightcyan}
Improvement & +0.16 & +0.81 & +0.62 & +6.86 & +6.95 & +6.06 & +9.66 & +11.04 & +10.09 \\
\hhline{|=|=|=|=|=|=|=|=|=|=|}
PointPillars & 88.52 & 79.29 & 76.34 & 57.27 & 51.00 & 46.44 & 83.88 & 62.77 & 59.50 \\
\hline
PointPillars+FasterRCNN & 89.52 & 80.11 & 77.14 & 70.38 & 63.98 & 58.13  & 88.07 & 73.88 & 69.07 \\
\hline
\rowcolor{lightcyan}
Improvement & +1.00 & +0.82 & +0.80 & +13.11 & +12.98 & +11.69 & +4.19 & +11.11 & +9.57 \\
\hhline{|=|=|=|=|=|=|=|=|=|=|}
PartA2 & 92.45 & 82.88 & 80.64 & 60.61 & 53.59 & 48.86 & 90.45 & 70.17 & 65.52 \\
\hline
PartA2+FasterRCNN & 92.98 & 83.80 & 81.37 & 72.44 & 65.52 & 58.98 & 94.01 & 79.39 & 74.28 \\
\hline
\rowcolor{lightcyan}
Improvement & +0.53 & +0.92 & +0.73 & +11.83 & +11.93 & +10.12 & +3.56 & +9.22 & +8.76 \\
\hhline{|=|=|=|=|=|=|=|=|=|=|}
PV-RCNN & 91.82 & 84.53 & 82.42 & 66.72 & 59.27 & 54.31 & 90.36 & 73.26 & 69.36 \\
\hline
PV-RCNN+FasterRCNN & 92.95 & 86.09 & 83.32 & 73.87 & 67.40 & 62.67 & 91.01 & 77.25 & 72.01 \\
\hline
\rowcolor{lightcyan}
Improvement & +1.13 & +1.56 & +0.90 & +7.15 & +8.13 & +8.36 & +0.65 & +3.99 & +2.65 \\
\hline
\end{tabular}
}
\label{table:single_mod_comp_3d}
\end{table}

\begin{table}[t]
\centering
\caption{Comparison with single modal detectors (BEV AP) on the KITTI val set.}
\resizebox{.92\textwidth}{!}{%
\begin{tabular}{|c|>{\centering\arraybackslash}p{1.2cm}|>{\centering\arraybackslash}p{1.2cm}|>{\centering\arraybackslash}p{1.2cm}|>{\centering\arraybackslash}p{1.2cm}|>{\centering\arraybackslash}p{1.2cm}|>{\centering\arraybackslash}p{1.2cm}|>{\centering\arraybackslash}p{1.2cm}|>{\centering\arraybackslash}p{1.2cm}|>{\centering\arraybackslash}p{1.2cm}|}
\hline
\multirow{2}{*}{Detector} & \multicolumn{3}{c|}{Car $AP_{BEV}$} & \multicolumn{3}{c|}{Pedestrian $AP_{BEV}$} & \multicolumn{3}{c|}{Cyclist $AP_{BEV}$} \\
\cline{2-10}
& Easy & Mod. & Hard & Easy & Mod. & Hard & Easy & Mod. & Hard \\
\hhline{|=|=|=|=|=|=|=|=|=|=|}
SECOND & 94.79 & 88.47 & 85.83 & 64.73 & 58.89 & 53.06 & 81.28 & 67.30 & 63.69 \\
\hline
SECOND+FasterRCNN & 95.75 & 89.69 & 87.05 & 73.01 & 66.93 & 60.49 & 91.33 & 80.30 & 75.37 \\
\hline
\rowcolor{lightcyan}
Improvement & +0.96 & +1.22 & +1.22 & +8.28 & +8.04 & +7.43 & +10.05 & +13.00 & +11.68 \\
\hhline{|=|=|=|=|=|=|=|=|=|=|}
PointPillars & 92.58 & 88.50 & 85.76 & 61.43 & 55.60 & 51.19 & 87.74 & 66.58 & 62.70 \\
\hline
PointPillars+FasterRCNN & 95.64 & 89.64 & 86.96 & 76.08 & 70.59 & 64.70 & 92.72 & 78.79 & 73.90 \\
\hline
\rowcolor{lightcyan}
Improvement & +3.06 & +1.14 & +1.20 & +14.65 & +14.99 & +13.51 & +4.98 & +12.21 & +11.20 \\
\hhline{|=|=|=|=|=|=|=|=|=|=|}
PartA2 & 93.55 & 89.38 & 87.13 & 64.19 & 58.05 & 52.22 & 93.87 & 73.46 & 68.83  \\
\hline
PartA2+FasterRCNN & 93.96 & 90.51 & 89.76 & 78.41 & 71.48 & 64.86 & 98.18 & 83.82 & 78.67 \\
\hline
\rowcolor{lightcyan}
Improvement & +0.41 & +1.13 & +2.63 & +14.22 & +13.43 & +12.64 & +4.31 & +10.36 & +9.84 \\
\hhline{|=|=|=|=|=|=|=|=|=|=|}
PV-RCNN & 94.43 & 90.78 & 88.67 & 69.53 & 62.12 & 57.18 & 92.81 & 75.55 & 70.88 \\
\hline
PV-RCNN+FasterRCNN & 95.92 & 92.63 & 90.07 & 77.65 & 72.70 & 68.03 & 94.93 & 80.30 & 75.43 \\
\hline
\rowcolor{lightcyan}
Improvement & +1.49 & +1.85 & +1.40 & +8.12 & +10.58 & +10.85 & +2.12 & +4.75 & +4.55 \\
\hline
\end{tabular}
}
\label{table:single_mod_comp_bev}
\end{table}

\begin{table}[t]
\centering
\caption{Performance comparison with multi-modal solutions on the KITTI val set.}
\resizebox{.9\textwidth}{!}{%
\begin{tabular}{|c|>{\centering\arraybackslash}p{1.2cm}|>{\centering\arraybackslash}p{1.2cm}|>{\centering\arraybackslash}p{1.2cm}|>{\centering\arraybackslash}p{1.2cm}|>{\centering\arraybackslash}p{1.2cm}|>{\centering\arraybackslash}p{1.2cm}|>{\centering\arraybackslash}p{1.2cm}|>{\centering\arraybackslash}p{1.2cm}|>{\centering\arraybackslash}p{1.2cm}|>{\centering\arraybackslash}p{1.2cm}|}
\hline
\multirow{2}{*}{Detector} & \multirow{2}{*}{\makecell{Speed \\ (FPS$^*$)}} & \multicolumn{3}{c|}{Car $AP_{3d}$} & \multicolumn{3}{c|}{Pedestrian $AP_{3d}$} & \multicolumn{3}{c|}{Cyclist $AP_{3d}$} \\
\cline{3-11}
& & Easy & Mod. & Hard & Easy & Mod. & Hard & Easy & Mod. & Hard \\
\hline
CLOCs-PVCas \cite{clocs_late_fusion} & - & 89.49 & 79.31 & 77.36 & 62.88 & 56.20 & 50.10 & 87.57 & 67.92 & 63.67 \\
\hline
Frustum PointNet \cite{frustum-pointnet} & 5.9 & 83.76 & 70.92 & 63.65 & 70.00 & 61.32 & 53.59 & 77.15 & 56.49 & 53.37 \\
\hline
Frustum PointPillars \cite{frustum_pointpillars} & 14.3 & 88.90 & 79.28 & 78.07 & 66.11 & 61.89 & 56.91 & 87.54 & 72.78 & 66.07 \\
\hline
PointPainting \cite{point-painting} & - & 88.38 & 77.74 & 76.76 & 69.38 & 61.67 & 54.58 & 85.21 & 71.62 & 66.98 \\
\hline
PointFusion \cite{point-fusion} & - & 77.92 & 63.00 & 53.27 & 33.36 & 28.04 & 23.38 & 49.34 & 29.42 & 26.98 \\
\hline
AVOD-FPN \cite{avod} & 10 & 84.41 & 74.44 & 68.65 & - & 58.80 & - & - & 49.70 & - \\
\hline
CAT-Det \cite{zhang2022catdetcontrastivelyaugmentedtransformer} & 10.2 & 90.12 & 81.46 & 79.15 & \textbf{74.08} & 66.35 & 58.92 & 87.64 & 72.82 & 68.20 \\
\hline
VirConv-T \cite{virconv} & 10.2 & \textbf{94.98} & \textbf{89.96} & \textbf{88.13} & 73.32 & 66.93 & 60.38 & 90.04 & 73.90 & 69.06 \\
\hline
LoGoNet \cite{li2023logonet} & - & 92.04 & 85.04 & 84.31 & 70.20 & 63.72 & 59.46 & \textbf{91.74} & 75.35 & \textbf{72.42} \\
\hline
MLF-DET-V \cite{lin2023mlf} & 10.8 & 89.70 & 87.31 & 79.34 & 71.15 & \textbf{68.50} & 61.72 & 86.05 & 72.14 & 65.42 \\
\hhline{|=|=|=|=|=|=|=|=|=|=|=|}
Ours (PointPillars+FasterRCNN) & 29.7 & 89.52 & 80.11 & 77.14 & 70.38 & 63.98 & 58.13  & 88.07 & 73.88 & 69.07 \\
\hline
Ours (PV-RCNN+FasterRCNN) & 10.1 & 92.95 & 86.09 & 83.32 & 73.87 & 67.40 & \textbf{62.67} & 91.01 & \textbf{77.25} & 72.01 \\
\hline
\end{tabular}
}
\label{table:multi_modal_comparison}
\end{table}

\subsection{Ablation Study} \label{sec:ablation}

We evaluate the contribution of each component of our module using the single-modal detector PointPillars as a baseline, a mainstream LiDAR detector in real-time applications. \cref{table:ablation} summarizes the results, where the overall AP is reported for both 3D and BEV, aggregated w.r.t. the difficulty of the detections. The advantages of incorporating RGB information can be seen already from the Bbox Matching module, which improves significantly the metrics by reducing the False Positive detections. Moreover, the Detection Recovery module provides further improvements, especially for moderate and hard cases, characterized by more False Negatives. This means that the method successfully recovers missed detections. Thus, the RGB detector finds objects that the LiDAR detector cannot detect. Finally, the Semantic Fusion module also contributes to the overall performance improvement, which confirms our hypothesis that the RGB branch is more reliable in providing semantic information. 

\textbf{Inference Speed}. We measure the inference speed of the proposed solution for a real-time application on one A100 GPU. While PointPillars is known to have a fast point cloud encoder, the Faster RCNN that we used provides a lower computational speed, around 37 FPS. As reported in \cref{table:ablation}, the computational overhead given by the three modules is low, allowing us to match real-time requirements, being modern LiDAR sensors' frame rates usually between 10 and 20 FPS. We measure the inference speed of each module separately, considering that, in a real-time application, the LiDAR and RGB branches can be parallelized. Thus, we do not sum the computational time of the LiDAR and RGB branches, but we take the slowest one.

\begin{table}[t]
\centering
\caption{Ablation studies on the KITTI val set.}
\resizebox{.7\textwidth}{!}{%
\begin{tabular}{|c|>{\centering\arraybackslash}p{1.1cm}|>{\centering\arraybackslash}p{1.1cm}|>{\centering\arraybackslash}p{1.1cm}|>{\centering\arraybackslash}p{1.1cm}|>{\centering\arraybackslash}p{1.1cm}|>{\centering\arraybackslash}p{1.1cm}|>{\centering\arraybackslash}p{1.1cm}|}
\hline
\multirow{2}{*}{Detector} & \multicolumn{3}{c|}{Overall $AP_{3d}$} & \multicolumn{3}{c|}{Overall $AP_{BEV}$} & \multirow{2}{*}{\makecell{Speed \\ (FPS)}}\\
\cline{2-7}
& Easy & Mod. & Hard & Easy & Mod. & Hard &\\
\hhline{|=|=|=|=|=|=|=|=|}
PointPillars & 76.56 & 64.35 & 60.77 & 80.59 & 70.23 & 66.55 & 62.5\\
\hline
FasterRCNN & - & - & - & - & - & - & 37.1\\
\hhline{|=|=|=|=|=|=|=|=|}
+ Bbox Matching & 80.87 & 69.94 & 65.42 & 86.32 & 76.43 & 72.48 & 35.4 \\
\hline
+ Detection Recovery & 81.05 & 71.92 & 67.57 & 86.57 & 78.45 & 74.50 & 29.7 \\
\hline
+ Semantic Fusion & \textbf{82.65} & \textbf{72.66} & \textbf{68.12} & \textbf{88.15} & \textbf{79.68} & \textbf{75.19} & 29.7 \\
\hline
\end{tabular}
}
\label{table:ablation}
\end{table}

\textbf{Effect of Frustum Proposals enlargement}. \cref{table:frustum-enlarge} shows the performance of PointPillars in the LiDAR branch for several enlargement factors for Cyclists and Pedestrians, which are the main categories interested by the  Detection Recovery module. It can be noticed how slightly enlarging the 2D bounding boxes is beneficial, especially for Cyclists. As the Frustum Localizer is trained to localize objects whose center is near the back-projection of the 2D bounding box center, increasing the enlargement factor too much does not result in significant performance degradation. However, it has to be noted that the number of points increases, and so does the computational complexity. Thus, we set the enlargement factor to 5\%.

\begin{table}[t]
\centering
\caption{Effect of the enlargement factor.}
\resizebox{.575\textwidth}{!}{%
\begin{tabular}{|>{\centering\arraybackslash}p{1.5cm}|>{\centering\arraybackslash}p{1.1cm}|>{\centering\arraybackslash}p{1.1cm}|>{\centering\arraybackslash}p{1.1cm}|>{\centering\arraybackslash}p{1.1cm}|>{\centering\arraybackslash}p{1.1cm}|>{\centering\arraybackslash}p{1.1cm}|}
\hline
\multirow{2}{*}{Enlarge \%} & \multicolumn{3}{c|}{Cyclist $AP_{3d}$} & \multicolumn{3}{c|}{Pedestrian $AP_{3d}$} \\
\cline{2-7}
& Easy & Mod. & Hard & Easy & Mod. & Hard \\
\hline
0\% & 86.36 & 72.52 & 67.76 & 70.37 & 63.90 & 58.02 \\
\hline
5\% & \textbf{88.07} & \textbf{73.88} & \textbf{69.07} & 70.38 & 63.98 & 58.13 \\
\hline
10\% & 86.29 & 73.82 & 69.05 & 70.50 & 64.02 & 58.13 \\
\hline
20\% & 86.28 & 73.79 & 67.71 & \textbf{70.67} & \textbf{64.09} & 58.12 \\
\hline
30\% & 86.30 & 72.42 & 67.65 & \textbf{70.67} & 64.06 & \textbf{58.24} \\
\hline
50\% & 86.24 & 72.49 & 67.64 & 70.60 & 63.99 & 58.08 \\
\hline
\end{tabular}
}
\label{table:frustum-enlarge}
\end{table}

\textbf{Effect of the Frustum Localizer}. In \cref{table:frustum-detector} we compare the performance of using the Frustum Localizer on Frustum Proposals with a simple baseline, based only on geometry and RGB information, showing how the Frustum Localizer performs better for both Cyclists and Pedestrians. The baseline sets the BEV center of the bounding boxes as the mean BEV coordinates between the two intersections of the two bottom lines of each frustum (projected onto the BEV). Predefined anchor sizes are used as dimensions of the bounding box. As it is not possible to provide an accurate estimation of the yaw angle, we greedily compare the width and the length of the anchor with the distribution of the BEV points between the maximum and minimum depth of the two intersections: we set it to 0 if the difference between the maximum and minimum coordinate in the x-axis is higher than the one on the y-axis, and to $\frac{\pi}{2}$ if it is lower. The center height coordinate is set as the mean of the points inside the vertical column corresponding to the same BEV bounding box. The orientation estimation is the main issue with the simple baseline together with the center estimation if the 2D bounding boxes are not accurate, justifying the additional computational effort for the Frustum Localizer.

\begin{table}[t]
\centering
\caption{Effect of the detector on the Frustum Proposals.}
\resizebox{.62\textwidth}{!}{%
\begin{tabular}{|c|>{\centering\arraybackslash}p{1.1cm}|>{\centering\arraybackslash}p{1.1cm}|>{\centering\arraybackslash}p{1.1cm}|>{\centering\arraybackslash}p{1.1cm}|>{\centering\arraybackslash}p{1.1cm}|>{\centering\arraybackslash}p{1.1cm}|}
\hline
\multirow{2}{*}{\makecell{Detector}} & \multicolumn{3}{c|}{Cyclist $AP_{3d}$} & \multicolumn{3}{c|}{Pedestrian $AP_{3d}$} \\
\cline{2-7}
& Easy & Mod. & Hard & Easy & Mod. & Hard \\
\hline
RGB Baseline & 86.35 & 70.19 & 65.45 & 67.83 & 60.83 & 54.62 \\
\hline
Frustum PointNet & \textbf{88.07} & \textbf{73.88} & \textbf{69.07} & \textbf{70.38} & \textbf{63.98} & \textbf{58.13}\\
\hline
\end{tabular}
}
\label{table:frustum-detector}
\end{table}

\section{Conclusions}

In this paper, we have proposed a hybrid late-cascade fusion approach that exploits a 3D LiDAR detector, a 2D RGB detector and the geometrical constraints of a stereo camera system. Our Detection Recovery module leverages RGB information to recover missed LiDAR detections. Our solution increases the performance of single-modal LiDAR detectors, especially for more challenging classes like Cyclists and Pedestrians. Moreover, our solution can combine any state-of-the-art detector (potentially without the need of re-training), without incurring in a prohibitive computational overhead.

\subsubsection{\ackname}
This paper is supported by the FAIR (Future Artificial Intelligence Research) project, funded by the NextGenerationEU program within the PNRR-PE-AI scheme (M4C2, Investment 1.3, Line on Artificial Intelligence) and by GEOPRIDE ID: 2022245ZYB, CUP: D53D23008370001 (PRIN 2022 M4.C2.1.1 Investment).
Model training and testing were possible thanks to the HPC grant from by the Ministry of Education, Youth and Sports of the Czech Republic through the e-INFRA CZ (ID:90254).

\bibliographystyle{splncs04}
\bibliography{bibliography}
\end{document}